\documentclass[a4paper,conference]{IEEEtran}
\IEEEoverridecommandlockouts
\usepackage{cite}
\usepackage{amsmath,amssymb,amsfonts}
\usepackage{algorithmic}
\usepackage{graphicx}
\usepackage{textcomp}
\usepackage{xcolor}
\def\BibTeX{{\rm B\kern-.05em{\sc i\kern-.025em b}\kern-.08em
    T\kern-.1667em\lower.7ex\hbox{E}\kern-.125emX}}
\usepackage{ragged2e}
\usepackage{ctable} 

\usepackage{epsfig}
\usepackage{amsmath}
\usepackage{amssymb}
\usepackage[ruled]{algorithm2e}
\usepackage{gensymb}
\DeclareMathOperator*{\argmin}{arg\,min}

\usepackage{hyperref}
\usepackage{multirow}
\usepackage{booktabs}

\usepackage{fancyhdr}
\fancypagestyle{firstpage}{%
  \lhead{25th International Conference on Pattern Recognition, ICPR 2020}
  \rhead{}
}

\begin{document}
\fancypagestyle{firstpage}{
  \lhead{25th International Conference on Pattern Recognition, ICPR 2020}
  \rhead{}
}
\title{Multi-Modal Deep Clustering: Unsupervised Partitioning of Images}

\author{\IEEEauthorblockN{Guy Shiran}
\IEEEauthorblockA{\textit{School of Computer Science and Engineering} \\
\textit{Hebrew University of Jerusalem}\\
Jerusalem, Israel \\
guy.shiran@mail.huji.ac.il}
\and
\IEEEauthorblockN{Daphna Weinshall}
\IEEEauthorblockA{\textit{School of Computer Science and Engineering} \\
\textit{Hebrew University of Jerusalem}\\
Jerusalem, Israel \\
daphna@cs.huji.ac.il}
}

\maketitle
\thispagestyle{firstpage}

\begin{abstract}
   The clustering of unlabeled raw images is a daunting task, which has recently been approached with some success by deep learning methods. Here we propose an unsupervised clustering framework, which learns a deep neural network in an end-to-end fashion, providing direct cluster assignments of images without additional processing. Multi-Modal Deep Clustering (MMDC), trains a deep network to align its image embeddings with target points sampled from a Gaussian Mixture Model distribution. The cluster assignments are then determined by mixture component association of image embeddings. Simultaneously, the same deep network is trained to solve an additional self-supervised task of predicting image rotations. This pushes the network to learn more meaningful image representations that facilitate a better clustering. Experimental results show that MMDC achieves or exceeds state-of-the-art performance on six challenging benchmarks. On natural image datasets we improve on previous results with significant margins of up to 20\% absolute accuracy points, yielding an accuracy of 82\% on CIFAR-10, 45\% on CIFAR-100 and 69\% on STL-10.
\end{abstract}

\section{Introduction}
\begin{figure}[t]
\begin{center}
   \includegraphics[width=1.\linewidth]{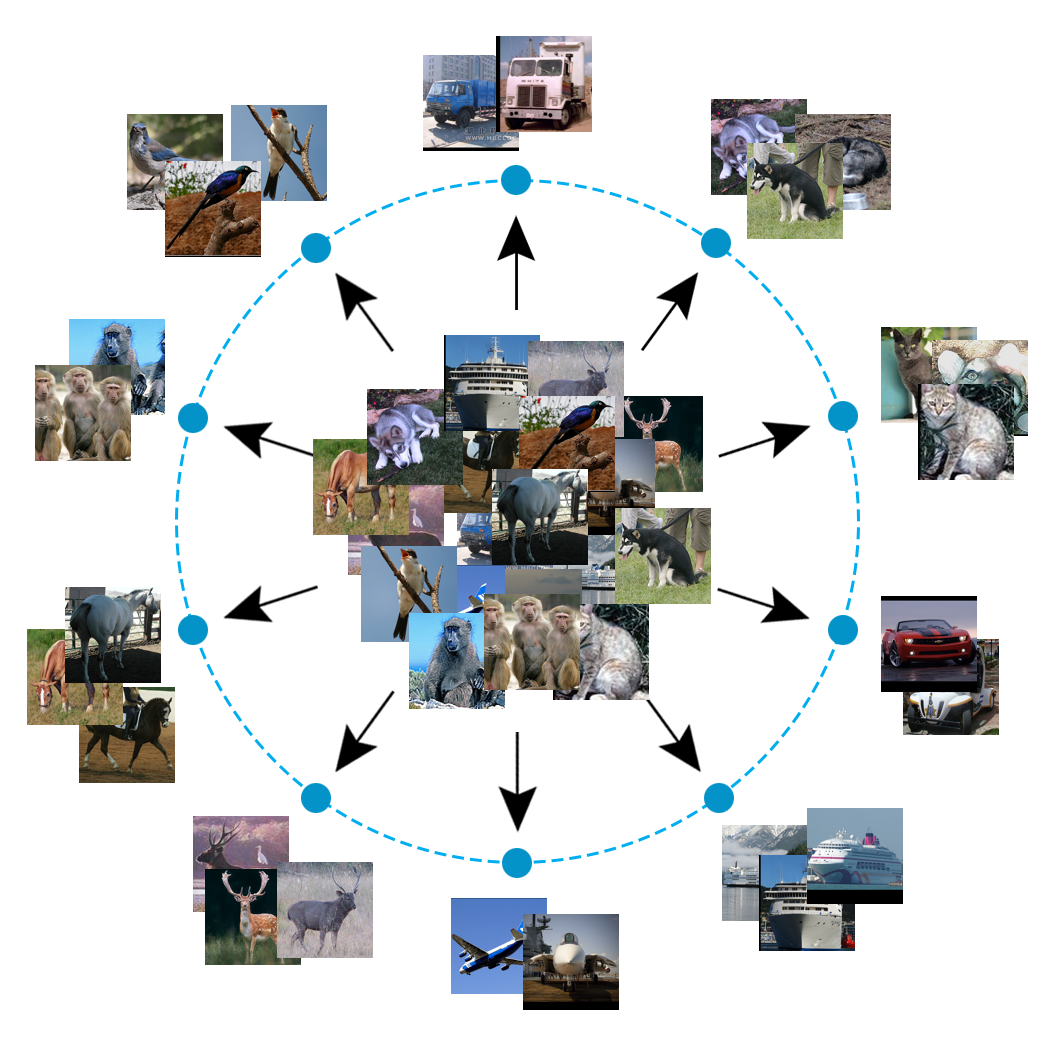}
   \vspace{-.3in}
\end{center}
   \caption{Our algorithm partitions a set of images into $k$ clusters by aligning image embeddings with target points sampled from a Gaussian Mixture Model on the $k$-dimensional unit sphere. }
   \vspace{-.142in}
\label{fig:mmnat_vis}
\end{figure}

Clustering involves the organization of data in an unsupervised manner, based on the distribution of datapoints and the distances between them. Since these properties are closely tied to the representation of the data, the problems of clustering and data representation are firmly connected and are therefore sometimes solved jointly. In accordance, in this work we start from a recent method for the unsupervised computation of effective data representation (or features discovery), and develop a clustering method whose results significantly improve the state of the art in the clustering of natural images. The method is illustrated in Fig~\ref{fig:mmnat_vis}. 

The task of unsupervised image clustering is challenging and interesting, as the algorithm needs to discover patterns in highly entangled data, and produce separated groups without explicitly specifying the grouping features. A large body of work has been devoted to the problem of clustering \cite{jain1999data}, see Section~\ref{sec:related} for a brief review of some recent related work. In recent years, with the emergence of deep learning as the method of choice in visual object recognition and image classification, emphasis has shifted to the computation of effective representations that will support successful clustering \cite{min2018survey}. Vice versa, unsupervised clustering loss has been used to drive the computation of image representation and the discovery of enhanced image features by making it possible to use unsupervised data in the training of deep networks, which traditionally require massive amounts of labeled data.

When learning feature representation from unsupervised data by minimizing a clustering-based loss function, one danger is cluster collapse - the representation may collapse to the trivial solution of a single cluster. In \cite{BJ2017}, a similar problem of representation collapse is managed by mapping the network's representation to a fixed set of randomly chosen points in some target features space. Here we borrow this mapping idea, and incorporate it into a clustering algorithm. 

More specifically, we first sample a fixed set of points in some target space. Since our method is designed to partition the data into $k$ clusters, 
the target points are chosen from a matched density function - Gaussian Mixture Model (GMM) with $k$ components. Our model trains a randomly initialized neural network to align its image embeddings with the sampled target points, directly inducing a partition that is based on the mixture components. This is done by simultaneously learning a one-to-one mapping between the output of the network and the target points, and updating the networks parameters to best fit images with their target points as assigned by the mapping.

In the absence of ground truth, the proposed approach is prone to instability as target points are continuously reassigned between images, creating a non-stationary online learning environment. Such instability is often linked with unsupervised learning tasks. To alleviate this problem, unsupervised tasks such as representation learning may be combined with self-supervision tasks to achieve better results \cite{doersch2017multi}. Here we adopt the approach taken by \cite{chen2018ssgan} to deal with the notorious instability of training generative adversarial networks. Thus the model is jointly trained on the main clustering task and on a self-supervised auxiliary task as defined in RotNet \cite{gidaris2018unsupervised}, where all images are subjected to 4 rotation angles. In this auxiliary task the network is trained to recognize the $2D$ rotation of each rotated image. 

For computation engine, our method uses off the shelf ConvNets and standard SGD training with mini-batch sampling in an end-to-end fashion. It is therefore scalable to large datasets. We evaluate our method on several standard benchmarks in image clustering, which is the goal of our method, significantly exceeding the state of the art on the 5 natural image datasets.

The rest of the paper is organized as follows: In Section~\ref{sec:related} we briefly review recent related work. In Section~\ref{sec:method} we describe our method in detail and elaborate on its various ingredients. Experimental results are reported in Section~\ref{sec:exp}.

\begin{figure*}[t]
\begin{center}
    \includegraphics[width=0.8\linewidth]{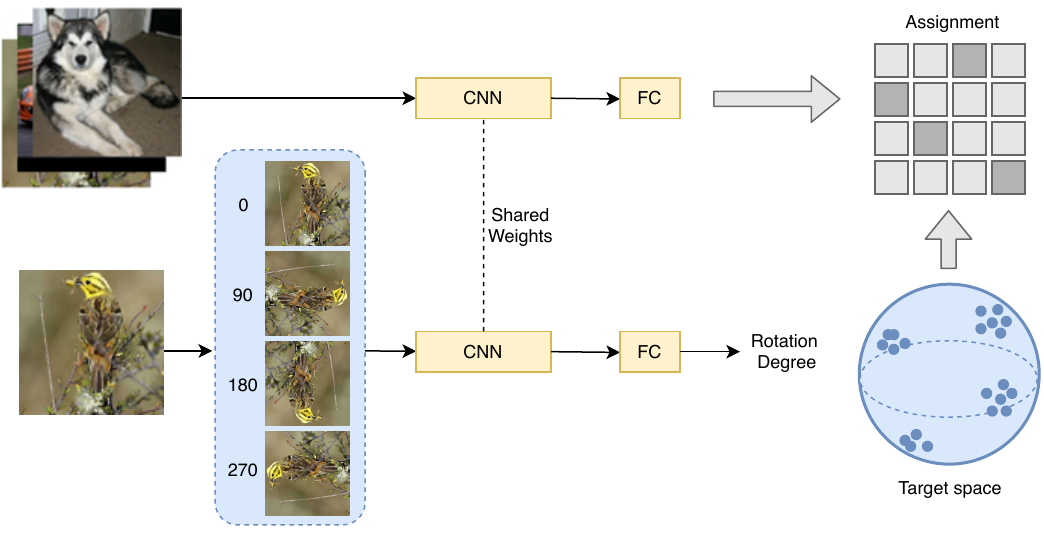}
\end{center}
   \caption{Our approach takes a set of images and solves two tasks in alternating epochs. In the primary task, a CNN is trained to produce output which matches some predefined set of target points sampled from a Gaussian mixture model, and optimally aligned with the training set. In the secondary task, given a rotated image, the same CNN is trained to predict the rotation angle of the image. }
\label{fig:short}
\end{figure*}

\section{Related work}
\label{sec:related}

\textbf{Data clustering.}
The objective of data clustering is to partition data points into groups such that points in each group are more similar to each other than to data points in the other groups. Traditionally, clustering methods have been divided into density-based methods \cite{kriegel2011density}, partition-based methods \cite{gerlhof1993partition}, and hierarchical methods \cite{duda1973pattern}. Partition-based methods, such as the popular k-means \cite{4031353,kmeansplus}, minimize a given clustering criterion by iteratively relocating data points between clusters until a (locally) optimal partition is attained. Density-based methods define clusters as areas with high density of points, separated by areas with low density of points \cite{DBSCAN}. Hierarchical based methods build a hierarchy of clusters in a top-to-bottom \cite{roux2015comparative} or bottom-to-top \cite{Gowda1978AgglomerativeCU} manner to determine clustering.

\textbf{Representation Learning.} Na\"ively attempting to cluster images with traditional approaches does not produce a pleasing partitions of the images, as they work on the raw representations of the images in pixel space, whereas semantically similar images are not necessarily similar in the high-dimensional pixel space in which the images reside. In recent years learning useful image representations in an unsupervised manner has been dominated by deep-learning-based approaches. Autoencoders (AEs) \cite{bengio2007} encode images with a deep network and are trained by reconstructing the image using a decoder network. These include several variations such as sparse AEs, denoising AEs \cite{denoiseae}, and more \cite{Masci2011StackedCA,SWWAE}. Generative models such as Generative Adversarial Networks (GAN) \cite{Goodfellow2014} and variational autoencoders (VAE) \cite{aevb} learn representations as a byproduct of learning to generate images. Tightly connected to our work, Noise-As-Targets (NAT) \cite{BJ2017} and DeepCluster \cite{caron2018deep} adopt a training strategy of iteratively reassigning psuedo-labels to points while training the network to fit them (see Section~\ref{sec:method}).

\textbf{Self-supervised learning.} A family of unsupervised learning algorithms that gained popularity in recent years are self-supervised methods. They learn representations by training a deep network to solve a pretext task, where labels can be produced directly from the data. Such tasks can be jigsaw puzzle solving \cite{Noroozi2016UnsupervisedLO}, predicting the relative position of patches in an image \cite{Doersch2015UnsupervisedVR}, generating image regions conditioned on their surroundings \cite{inpaintpathak2016}, or more recently predicting image rotations (RotNet) \cite{gidaris2018unsupervised}. In self-supervised GANs \cite{chen2018ssgan}, predicting image rotations is used as an auxiliary task to stabilize and improve training, by enhancing the discriminator's representation capabilities. Here we adopt this approach as well, as elaborated later on.

\textbf{Deep clustering.} The dominant and most successful approach to clustering of images in recent years has been to incorporate the tasks of representation learning and clustering into a single framework. Prominent works in the past years have been Joint Unsupervised Learning (JULE) \cite{yangCVPR2016joint}, where the authors adopt an agglomerative clustering approach by iteratively merging clusters of deep representations and updating the networks parameters. Deep Adaptive Clustering (DAC) \cite{Chang2017DeepAI} recasts the clustering problem into a binary pairwise-classification framework, where cosine distances between image features of image pairs are used as a similarity measure to decide if they belong to the same cluster. Associative Deep Clustering (ADC) \cite{Husser2017AssociativeDC} jointly learns network parameters and embedding centroids with an association loss in order to estimate cluster membership. More recently, Invariant Information Clustering (IIC) \cite{iic} adopts an approach that achieves clustering based on maximizing the mutual information between two sets: deep embeddings of images, and instances of the images that underwent  random image transformations while keeping the image semantic meaning intact. IIC leverages auxiliary over-clustering to increase expressivity in the learned feature representation, improving the representation capabilities of its network. This tactic bears resemblance to our incorporation of rotation prediction as an auxiliary task.

\section{Method}
\label{sec:method}

Our goal is to partition a set of images into $k$ clusters, which reflect internal structure in the data. Fig.~\ref{fig:short} shows an overview of the proposed approach. The algorithm alternates between solving the main unsupervised clustering task, and an auxiliary self-supervised task that helps the training process. The ingredients of the method are described next. The full method is summarized in Algorithm~\ref{algo:1}.

\subsection{Unsupervised learning}

The starting point for this work is an unsupervised learning framework for learning image representation from unlabeled data. The method, Noise as Targets \cite{BJ2017}, learns useful representations of images by training a deep network to align its images' embeddings with a fixed set of target points. The target points are uniformly scattered on the $d$-dimensional unit sphere. 

More specifically, let $X=\{x_i\}_{i=1}^n$ denote a set of images, and $f_\theta:X \rightarrow Z$ the parameterized deep network we wish to train. The output of $f_\theta$ is normalized to have $\ell_2$ norm of $1$, entailing that $Z$ is the $d$-dimensional unit sphere. NAT starts by uniformly sampling $n$ targets on this unit sphere. Let $\{t_i\}_{i=1}^n$ denote the set of target points, which remain fixed throughout the training. Each image $x_i$ is assigned a unique target $y_i$ through a permutation $P:[n]\rightarrow [n]$. The optimization objective is formulated as
\begin{align}\label{eq:1}
\min\limits_{\theta, P} \frac{1}{n} \sum_{i} \ell(f_\theta (x_i), y_i) ~~~~~~
y_i=t_{P(i)}
\end{align}
where $\ell$ is the Euclidean distance.

This optimization problem is solved in a stochastic manner, by iteratively solving it over randomly sampled mini-batches. Given a mini-batch of images $X_b$, the current representation vectors $f_\theta(X_b)$ are first computed. Subsequently, Equation~(\ref{eq:1}) is optimized for $P$ over the points in mini-batch $X_b$ using the Hungarian method \cite{Kuhn55thehungarian}, which reassigns the currently assigned targets of the mini-batch to minimize the Euclidean distance ($\ell_2$) between images and their assigned target points. Finally, the gradients of $f_\theta$ on $X_b$ with respect to $\theta$ are computed, and an SGD step is executed.

Intuitively, NAT permutes the assignment of image representation vectors to target points delivered by $f_\theta$, so that nearby embedding vectors are mapped to nearby target vectors, and then updates $\theta$ accordingly. This process leads to the grouping  of semantically similar images in target space, and to effective representations that perform well in downstream computer vision classification and detection tasks.

\begin{algorithm}[bt]
\caption{}
\label{algo:1}
\begin{algorithmic}
   \STATE {\bfseries Input:} \newline
       $\{x_i\}^{n}_{i=1}$ - images \newline
       $f_\theta$ - ConvNet with two heads \newline
       $k$ - number of clusters \newline
       $epochs$ - number of epochs to train \newline
       $iters$ - number of iterations in an epoch \newline
       $\sigma$ - variance of normal distribution \newline
       $d$ - dimension of embedding space \newline
       $\lambda_c, \lambda_r$ - learning rates \newline
       $g$ - random image transformation \newline
       $r$ - number of instances of $g$ in a batch
   \STATE {\bfseries Init:} \newline
       $P\leftarrow$  initialize with random assignments \\
       $\theta \leftarrow$ initialize with random weights \\
       $T \leftarrow$ initialize empty list \\
    \FOR {$i=1...n$}
    \STATE sample $c \sim Categ(\frac{1}{K},...,\frac{1}{K})$ \\
    \STATE sample $u \sim N(\mu_c,\sigma \cdot I_{d \times d})$\\
    \STATE $T[i] \leftarrow t_i=\frac{u}{\|u\|}$
    \ENDFOR
   
   \FOR {$e=1...epochs$}
    \FOR {$i=1...iters$}
       \STATE sample batch $X_b$ and assigned targets $T_b$
       \STATE compute $f_\theta (X_b)$
       \STATE update $P$ by minimizing Equation~(\ref{eq:1}) w.r.t $P$
       \STATE compute $\nabla_\theta L_c(\theta)$ of Equation~(\ref{eq:1}) for $g(X_b)$
       \STATE update $\theta \leftarrow \theta - \lambda_c \nabla_\theta L_c(\theta)$
     \ENDFOR
     
    \FOR {$i=1...iters$}
       \STATE sample batch $X_b$
       \STATE rotate $X_b$  $\forall r\in \{ 0^{\degree}, 90^{\degree}, 180^{\degree}, 270^{\degree}\}$
       \STATE compute $\nabla_\theta L_r(\theta)$  // $L_r$ is cross-entropy loss
       \STATE update $\theta \leftarrow \theta - \lambda_r \nabla_\theta L_r(\theta)$
       
     \ENDFOR
     \ENDFOR
\end{algorithmic}
\end{algorithm}

\subsection{Multi-modal distribution of target points}

The uniform distribution of target points on the unit sphere, as described above, is not well suited for unsupervised clustering, since it is likely to blur the dividing lines between clusters rather than sharpen them. Instead,  multi-modal distribution seems like a natural choice for the objective of clustering, as it directly produces separated groups in target space. 

In this work, we propose to use the mixture of Gaussians distribution, projected to the unit sphere, for the sampling of target points. Formally, this implies:
\begin{equation}
\begin{split}
\label{eq:def-gmm}
p (u) &= \sum^{K}_{k=1} \alpha_k \cdot p_k (u) ~~~~~~~~~~~~u\in {\mathbb R}^d \\
p (t_i) &= \int_{\frac{u}{\|u\|_2}=t_i} p (u)du ~~~~~~~~~t_i\in Z
\end{split}
\end{equation}
where $K$ denotes the number of Gaussians in the mixture, $d$ the dimension of the embedding space, $\alpha_{k=1..K}$  a categorical random variable, and $p_k(u)$  the multivariate normal distribution $N(\mu_k,\Sigma_k)$, parameterized by mean vector $\mu_k$ and covariance matrix $\Sigma_k$. In the absence of prior knowledge we assume that the mixture components are equally likely, namely $\alpha_k=\frac{1}{K}\forall k\in [K]$. Finally, since the target points are constrained to lie on the unit sphere, we project the sample in $\mathbb{R}^d$ to the unit sphere by $t_i=\frac{u}{\|u\|_2}$. 

We define the cluster assignment $c_i$ of image $x_i$ as follows
\begin{align}\label{eq:2}
c_i = \argmin\limits_k \|f_\theta(x_i)-\mu_k \|_2
\end{align}
Note that if the final network $f_\theta$ fits that target points exactly, namely $f_\theta(x_i)=y_i$, and if $\Sigma_k$ are the same $\forall k$, then with high probability $c_i$ is the index of the mixture component from which target point $y_i$ has been sampled.

\subsection{Image Transformations}
Data augmentation is a useful and common technique to improve performance of machine learning algorithms. Usually, random image transformations such as cropping, flipping, rotation, scaling and photometric transformations are applied to images in order to expand the dataset with new and unique images. In our task of unsupervised clustering, these random transformations are essential, because they provide several instances of the same image that appear different but share the same semantic meaning as they contain the same object. Let $g$ denote a random image transformation. In our method, we use the center crop of an image when minimizing Equation~(\ref{eq:1}) w.r.t $P$. When minimizing the same equation w.r.t $\theta$, we first apply $g$ to the image. Why is this algorithmic ingredient useful? When training the ConvNet, it must find common patterns between the original images and transformed images when fitting them to the same target. These common patterns are likely to appear in other images in the dataset belonging to the same class. This pushes the network to map images that contain the same objects closer to each other, in a similar manner to the beneficial effect of self-supervision.

\begin{table*}[t]
\small
\caption{\protect\justify \textbf{Unsupervised clustering results.} The results of our method are shown below the separation line. For each dataset, we show the average result over five runs, standard error (ste) and the best run. Above the separation line we list state of the art results for comparison, see review in Section~\ref{sec:related}. Unreported results are marked with (-).}

\begin{center}
\begin{tabular}{l l c c c c c c c c c c c c}
\toprule
&  & \multicolumn{2}{c}{MNIST} & \multicolumn{2}{c}{CIFAR-10} & \multicolumn{2}{c}{CIFAR-100} & \multicolumn{2}{c}{STL-10}  & \multicolumn{2}{c}{ImageNet-10} & \multicolumn{2}{c}{Tiny-ImageNet} \\
& & NMI & ACC & NMI & ACC & NMI & ACC & NMI & ACC & NMI & ACC & NMI & ACC \\
\midrule
\multicolumn{2}{l}{k-means} & 0.499 & 0.572 & 0.087 & 0.228 & 0.083 & 0.129 & 0.124 & 0.192 & 0.119 & 0.241 & 0.065 & 0.025 \\
\multicolumn{2}{l}{SC} & 0.663 & 0.696 & 0.103 & 0.247 & 0.090 & 0.136 & 0.098 & 0.159 & 0.151 & 0.274 & 0.063 & 0.022 \\
\multicolumn{2}{l}{AE} & 0.725 & 0.812 & 0.239 & 0.313 & 0.100 & 0.164 & 0.249 & 0.303 & 0.210 & 0.317 & 0.131 & 0.041 \\
\multicolumn{2}{l}{DEC (2016)} & 0.772 & 0.843 & 0.257 & 0.301 & 0.136 & 0.185 & 0.276 & 0.359 & 0.282 & 0.381 & 0.115 & 0.037 \\
\multicolumn{2}{l}{JULE (2016)} & 0.913 & 0.964 & 0.192 & 0.272 & 0.103 & 0.137 & 0.182 & 0.277 & 0.175 & 0.300 & 0.102 & 0.033 \\
\multicolumn{2}{l}{DAC (2017)} & 0.935 & 0.978 & 0.396 & 0.522 & 0.185 & 0.238 & 0.249 & 0.303 & 0.394 & 0.527 & 0.190 & 0.066 \\
\multicolumn{2}{l}{IIC (2019)}  & \textbf{0.978} & \textbf{0.992} & 0.513 & 0.617 & 0.224 & 0.257 & 0.431 & $\ $0.499$^1$ & - & - & - & - \\
\multicolumn{2}{l}{DCCM (2019)} & - & - & 0.496 & 0.623 & 0.285 & 0.327 & 0.376 & 0.482 & 0.608 & 0.710 & 0.224 & 0.108 \\
\midrule
\multirow{3}{*}{Ours} & avg. & 0.971 & 0.990 & \textbf{0.703} & \textbf{0.820} & \textbf{0.418} & \textbf{0.446} & \textbf{0.593} & \textbf{0.694} & \textbf{0.719} & \textbf{0.811} & \textbf{0.274} & \textbf{0.119} \\
& ste & $\pm$.000 & $\pm$.000 & $\pm$.011 & $\pm$.019 & $\pm$.003 & $\pm$.006 & $\pm$.005 & $\pm$.013 & $\pm$.008 & $\pm$.012 & $\pm$.001 & $\pm$.001 \\
& best & 0.973 & 0.991 & 0.720 & 0.843 & 0.423 & 0.464 & 0.609 & 0.741 & 0.732 & 0.830 & 0.277 & 0.121 \\
\bottomrule
\end{tabular}
\end{center}
\label{table:3}
\end{table*}

\begin{figure*}[t]
\begin{center}
   \includegraphics[width=1.\linewidth]{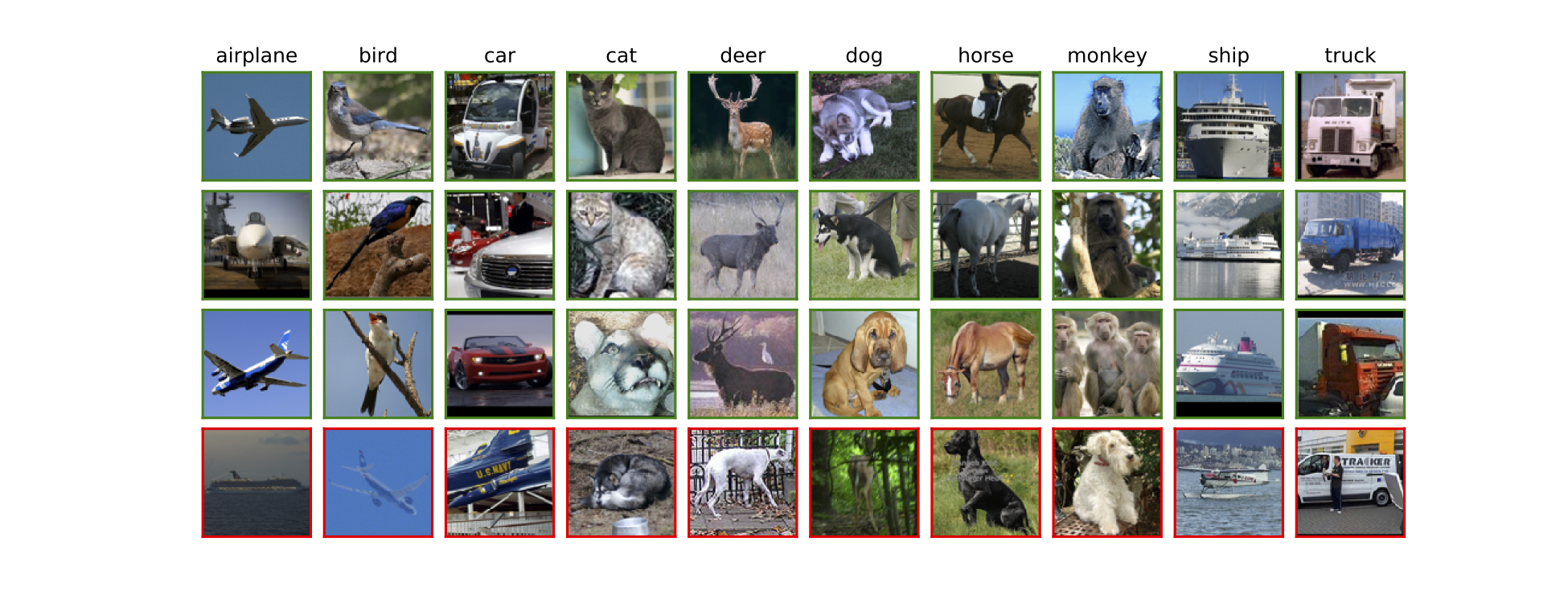}
   \vspace{-.6in}
\end{center}
   \caption{Unsupervised image clustering results on STL-10. Each column shows images from a different cluster. The top three images in each column are examples of images from the same class successfully clustered together. The images in the fourth row illustrate failure cases, where the image is assigned to the wrong cluster (e.g., an airplane assigned to the 'bird' cluster).}
\label{fig:stl-10}
\end{figure*}

\subsection{Auxiliary task}

While optimizing the clustering objective (\ref{eq:1}), the ConvNet model simultaneously learns image representation and partitions the images. The success of unsupervised clustering is highly correlated with the quality of the learnt representation. It has been repeatedly shown that self-supervision methods can significantly improve the quality of representations in an unsupervised learning scenario. To benefit from this idea, we employ RotNet \cite{gidaris2018unsupervised}, which is a self-supervised learning algorithm that learns image features by training a ConvNet to predict image rotations. Specifically, images are rotated by $r$ degrees where $r\in \{ 0^{\degree}, 90^{\degree}, 180^{\degree}, 270^{\degree}\}$, and the model is subsequently trained to predict their rotation by optimizing the cross-entropy loss. RotNet produces competitive performance in representation learning benchmarks, and has been shown to benefit training in other tasks, when incorporated into a model as an auxiliary task \cite{chen2018ssgan,Gidaris2019BoostingFV,lucic2019fewlabelsgans}. We incorporate RotNet into our method, modifying the ConvNet training procedure to alternate between optimizing the main clustering task and this secondary auxiliary task.

\subsection{Refinement Stage}

As we have no prior knowledge regarding the size of the clusters, we begin by assuming that clusters' sizes are equal. When this assumption cannot be justified, we propose to augment the algorithm with an additional step, performed after the main training is concluded. In this step the assumption is relaxed, while target points are iteratively reassigned based on the outcome of k-means applied to $f_\theta(x_1),...,f_\theta(x_n)$, and assigning image $x_i$ to target $\mu_j$ with label $j\in[K]$ derived from the outcome of k-means. This ingredient is similar to DeepCluster \cite{caron2018deep}, proposed by Caron et al. as an approach for representation learning, where they perform the clustering on the latent vectors of the model and not the final output layer. A possible alternative method may start with this stage and discard the first one altogether, as this approach makes no assumption on the size of the clusters. However, we found that starting off with reassigning labels based on k-means is not competitive and produces less accurate clusters. For example, training on MNIST results in low accuracy of $81\%$ ($\pm 2.67$).

\section{Experiments}
\label{sec:exp}

We tested our method on several image datasets that are commonly used as benchmark for clustering, see results in Table~\ref{table:3}. We compare ourselves to state-of-the-art methods such as DEC \cite{Xie2015UnsupervisedDE}, JULE \cite{yangCVPR2016joint}, DAC \cite{Chang2017DeepAI}, IIC \cite{iic} and DCCM \cite{wu2019deep}. In almost all cases our method improves on previous results significantly\footnote{Note that with STL-10, IIC reports an accuracy of $0.596$ when using the much larger unlabeled data segment that includes distractor classes.}. Examples of clustering results on the STL-10 dataset of natural images are shown in Figure~\ref{fig:stl-10}.

\begin{figure*}[t]
\begin{center}
\includegraphics[width=1.\textwidth]{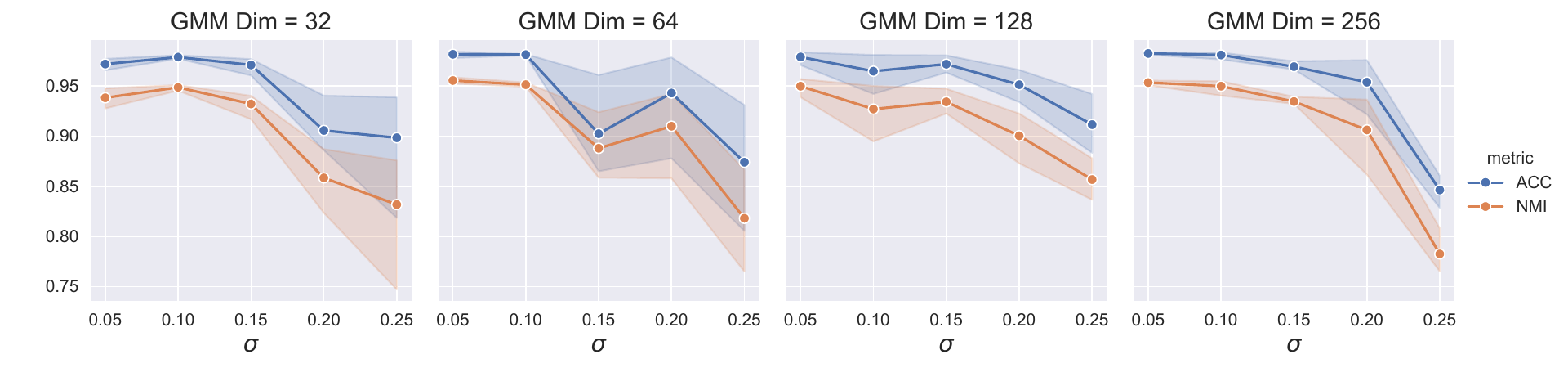}
\end{center}
\caption{Comparison of clustering performance on MNIST with different Mixture of Gaussians initializations. We compare different dimensions for the target vectors and different coefficient parameters ($\sigma$) for the covariance matrices of the gaussians. These results do not include performing the refinement stage.}
\label{fig:mnist_gmm}
\end{figure*}


In the rest of this section we specify the implementation details of our method, and analyze the results. Subsequently, we report the results of an ablation study evaluating the various ingredients of the algorithm, which demonstrate how they contribute to its success. Our code is available online\footnote{https://github.com/guysrn/mmdc}. 

\subsection{Implementation details and evaluation scores}
\label{sec:impl-det}

\textbf{Datasets.} Six datasets are used in our empirical study: MNIST \cite{Lecun98gradient-basedlearning}, CIFAR-10 \cite{cifar10}, the 20 superclasses of CIFAR-100 \cite{cifar10}, STL-10 \cite{stl10}, ImageNet-10 (a subset of ImageNet \cite{imagenet2009}) and Tiny-ImageNet \cite{imagenet2009}, see Table~\ref{table:datasets}. We are most interested in the datasets that consist of natural images. These datasets are commonly used to evaluate clustering methods.

\begin{table}[h]
\begin{center}
\caption{The image datasets used in our experiments.}
\label{table:datasets}
\begin{tabular}{l c c c}
\toprule
Name & Classes & Samples & Dimension \\
\midrule
MNIST & 10 & 70,000 & 28$\times$28 \\
CIFAR-10 & 10 & 60,000 & 32$\times$32$\times$3 \\
CIFAR-100 & 20 & 60,000 & 32$\times$32$\times$3 \\
STL-10 & 10 & 13,000 & 96$\times$96$\times$3 \\
ImageNet-10 & 10 & 13,000 & 96$\times$96$\times$3 \\
Tiny-ImageNet & 200 & 100,000 & 64$\times$64$\times$3 \\
\bottomrule
\end{tabular}
\end{center}
\end{table}

\textbf{Architectures.} 
For the MNIST experiments we use a small VGG model \cite{vgg} with batch normalization \cite{Ioffe2015BatchNA}. Each block in this neural network consists of one convolution layer, followed by a batch normalization layer and ReLU activation function, and ends with a max pooling layer. Our model has four blocks. For all other experiments we use a ResNet model \cite{2015DeepRL} with 18 layers. These base models are followed by a linear prediction layer, that outputs the cluster assignments. When trained on the auxiliary task, the base model is also followed by another linear head, which predicts the image rotation.

\textbf{Training details.} 
The network is trained with stochastic gradient descent with learning rate $0.05$ and momentum of $0.9$. We apply weight decay of $0.0001$ for CIFAR-100 and Tiny-ImageNet, and $0.0005$ for all other datasets. We use batch size $128$ and perform random image augmentations which include cropping, flipping and color jitter. When training on the auxiliary rotation task, we rotate each image to all four orientations, resulting in an effective batch size of $512$. We train the network for $400$ epochs and decay learning rate by a factor of $5$ after $350$ epochs. For MNIST we train for $50$ epochs and decay learning rate by a factor of $10$ after $40$ epochs. Training on CIFAR-10 takes 10.5 hours on a single GTX-1080 GPU.

\textbf{Mixture of Gaussians.} We examined several initialization heuristics to determine the Gaussian means $\{\mu_k\}$ in the GMM distribution defined in (\ref{eq:def-gmm}) and the covariance matrices $\{\Sigma_k\}$. A comparison of different initialization schemes is provided in Figure~\ref{fig:mnist_gmm}, where all vectors lie on the $d$-dimensional unit sphere. Gaussian means $\{\mu_k\}$ are sampled from a multi-variate uniform distribution within the range $[-0.1,0.1]$ and projected onto the unit sphere. We always set $\Sigma_k=\sigma \cdot I_{K \times K} \ \forall k \in [K]$. We compare different values for the dimension $d$ and the variance parameter $\sigma$. Smaller variance usually performs best with the added benefit of similar performance for different choices of dimension $d$. We therefore opted to use $K$ different one-hot vectors in $\mathbb{R}^K$ for $\{\mu_k\}$ with variance $\sigma=0$, as this achieved good performance while reducing the number of free hyperparameters.

\textbf{Evaluation scores.} 
To evaluate clustering performance we adopt two commonly used scores: Normalized Mutual Information (NMI), and Clustering Accuracy (ACC). Clustering accuracy measures the accuracy of
the hard-assignment to clusters, with respect to the best permutation of the dataset’s ground-truth labels. Normalized Mutual Information measures the mutual information between the ground-truth labels and the predicted labels based on the clustering method. The range of both scores is [0, 1], where a larger value indicates more precise clustering results. We use centrally cropped images for evaluation.

\subsection{Empirical Analysis}

The results of our method when applied to the six image datasets are reported in Table~\ref{table:3}. Clearly, our clustering algorithm is able to separate unlabeled images into distinct groups of semantically similar images with high accuracy, improving the state-of-the-art in the five datasets of natural images. Compared to previous state-of-the-art, we improve clustering accuracy on CIFAR-10 by 20\%, CIFAR-100 by 12\%, STL-10 by 20\%, ImageNet-10 by 10\% and Tiny-ImageNet by 1\%.

In the results reported in Table~\ref{table:3}, the refinement stage was invoked only when using the MNIST dataset. A more complete ablation study of the refinement stage is reported in Table~\ref{table:refine}. 
The auxiliary task of RotNet, which was shown to be beneficial when learning natural images, was used to enhance the clustering of all the datasets except MNIST. For reference, we used the same image augmentations as in \cite{iic}, which uses a larger ResNet-34 as the backbone for the model. 

\begin{table}[h]
\begin{center}
\caption{Clustering performance on CIFAR-10, showing the combined effect of pre-processing with the Sobel filter and adding a rotation loss. First row: no pre-processing and no rotation loss, second row: pre-processing and no rotation loss, third row: no pre-processing with a rotation loss, fourth row: both.}
\label{table:sobel_rotations}
\begin{tabular}{c c c c}
\toprule
Sobel & Rotation loss & NMI & ACC \\
\midrule
& & 0.428 $\pm$ .005 & 0.492 $\pm$ .003 \\
\checkmark & & 0.463 $\pm$ .003 & 0.560 $\pm$ .006   \\
& \checkmark & 0.703 $\pm$ .011 & 0.820 $\pm$ .019 \\
\checkmark & \checkmark & 0.610 $\pm$ .010 & 0.725 $\pm$ .020 \\
\bottomrule
\end{tabular}
\end{center}

\end{table}

\begin{table*}[t]
\begin{center}
\small
\caption{\protect\justify Evaluation of unsupervised feature learning methods on CIFAR-10 and CIFAR-100. We use the penultimate layer of the network as image features and test performance with two procedures. We perform k-means clustering on the image features and train a linear classifier using the image labels. As a reference, we report results using an imagenet-pretrained  ResNet-18.}
\label{table:nat_rotnet}
\begin{tabular}{l c c c c c c}
\toprule
& \multicolumn{3}{c}{CIFAR-10} & \multicolumn{3}{c}{CIFAR-100} \\
& \multicolumn{2}{c}{K-means} & Linear & \multicolumn{2}{c}{K-means} & Linear \\
& NMI & ACC & ACC & NMI & ACC & ACC \\
\midrule
ImageNet labels & 0.321 & 0.407 & 0.782 & 0.247 & 0.281 & 0.646 \\
\midrule
NAT & 0.044 $\pm$ .001 & 0.162 $\pm$ .001 & 0.315 $\pm$ .002 & 0.037 $\pm$ .001 & 0.095 $\pm$ .001 & 0.177 $\pm$ .001 \\
RotNet & 0.329 $\pm$ .011 & 0.349 $\pm$ .012 & 0.740 $\pm$ .002 & 0.261 $\pm$ .006 & 0.284 $\pm$ .013 & 0.543 $\pm$ .001 \\
NAT+RotNet & 0.413 $\pm$ .005 & \textbf{0.511 $\pm$ .002} & 0.764 $\pm$ .001 & 0.190 $\pm$ .007 & 0.232 $\pm$ .006 & 0.499 $\pm$ .002 \\
\midrule
Ours & \textbf{0.428 $\pm$ .011} & 0.397 $\pm$ .018 & \textbf{0.869 $\pm$ .002} & \textbf{0.395 $\pm$ .002} & \textbf{0.347 $\pm$ .007} & \textbf{0.662 $\pm$ .001} \\
\bottomrule
\end{tabular}

\end{center}
\end{table*}

\begin{table*}[t]
\begin{center}
\small
\caption{\protect\justify Comparison of clustering performance before and after the refinement stage.}
\label{table:refine}
\begin{tabular}{l l c c c c c c c c c c c c}
\toprule
 & & \multicolumn{2}{c}{MNIST} & \multicolumn{2}{c}{CIFAR-10} & \multicolumn{2}{c}{CIFAR-100} & \multicolumn{2}{c}{STL-10}  & \multicolumn{2}{c}{ImageNet-10} & \multicolumn{2}{c}{Tiny-ImageNet} \\
& & NMI & ACC & NMI & ACC & NMI & ACC & NMI & ACC & NMI & ACC & NMI & ACC \\
\midrule
\multirow{ 2}{*}{Before} & avg. & 0.950 & 0.981 & 0.703 & 0.820 & 0.418 & 0.446 & 0.593 & 0.694 & 0.719 & 0.811 & 0.274 & 0.119 \\
& ste & $\pm$.002 & $\pm$.001 & $\pm$.011 & $\pm$.019 & $\pm$.003 & $\pm$.006 & $\pm$.005 & $\pm$.013 & $\pm$.008 & $\pm$.012 & $\pm$.001 & $\pm$.001 \\
\midrule
\multirow{ 2}{*}{After} & avg. & 0.971 & 0.990 & 0.715 & 0.829 & 0.422 & 0.446 & 0.596 & 0.696 & 0.725 & 0.815 & 0.254 & 0.095 \\
& ste & $\pm$.000 & $\pm$.000 & $\pm$.009 & $\pm$.021 & $\pm$.002 & $\pm$.005 & $\pm$.005 & $\pm$.013 & $\pm$.008 & $\pm$.012 & $\pm$.001 & $\pm$.002 \\
\bottomrule
\end{tabular}
\end{center}
\end{table*}

\textbf{Benefits of auxiliary task.}
Applying the Sobel filter to an image emphasizes edges and discards colors. This pre-processing is commonly done in the context of unsupervised representation learning and clustering algorithms, presumably to avoid sub-optimal solutions based on trivial cues such as color \cite{BJ2017, iic}. We observed an interesting interaction between Sobel filtering and training with the auxiliary task of predicting image rotations. Without the auxiliary task, Sobel filtering indeed improves clustering performance as seen in Table~\ref{table:sobel_rotations}. In contrast, when training with an auxiliary task and adding the rotation loss, pre-processing with the Sobel filter degrades the algorithms performance. Furthermore, without the rotation loss the learning rate has to be reduced to $0.01$ for training to converge. The reason may be that trivial cues such as color are not beneficial for the task of predicting image rotations, and therefore the auxiliary task forces the ConvNet to learn features that focus on the object in the image. Once the focus is on the object, additional cues such as color can be beneficial for clustering, and as a result pre-processing with the Sobel filter is detrimental to the algorithm's performance.

\textbf{Feature Evaluation.}
Our algorithm borrows some of its ingredients from NAT and RotNet. However, while these two methods address representation learning, the final goal of our method is clustering. Nevertheless, we compare our method to NAT and RotNet in two ways. First, we examine the clustering capabilities of the methods by applying k-means to the penultimate layer of the networks. Second, we evaluate the learnt features by training a linear classifier with the image labels on top of the frozen features of the networks. We use the same architecture and image transformations as our model for both methods. We follow the training procedure from \cite{kolesnikov2019revisiting} for training RotNet and \cite{BJ2017} for training NAT. 

More specifically, we train the linear classifier with stochastic gradient descent with learning rate 0.1, momentum of 0.9, weight decay of 0.00001, batch size of 128, cosine annealing for learning rate scheduling, and 100 training epochs. Results with CIFAR-10 and CIFAR-100 are reported in Table~\ref{table:nat_rotnet}. As shown our method outperforms the others in all cases except one, where NAT+RotNet performs better when clustering CIFAR-10 image features. As a reference for the linear classifier performance, we also evaluate a model pretrained with ImageNet (first row in Table~\ref{table:nat_rotnet}). Note that we use the same image augmentations as for training the unsupervised methods, including 20$\times$20 cropping, which may degrade performance for this model.

\textbf{Refinement stage.} We compare clustering performance with and without the proposed refinement stage in Table~\ref{table:refine}. MNIST is the only dataset with class imbalance, as its smallest class has $6313$ samples while its largest has $7877$. Reassuringly, the refinement stage helps the algorithm achieve near perfect clustering with accuracy of 99.0\%.

\section{Summary}
For the task of unsupervised semantic image clustering, we presented an end-to-end deep clustering framework, that trains a ConvNet to align image embeddings with targets sampled from a Gaussian Mixture Model by solving a linear assignment problem using the Hungarian algorithm. To achieve effective training, we incorporated an additional auxiliary task - the prediction of image rotation. Our ablation study shows that the contribution of this component is essential for the success of the method. Even though the proposed method is quite simple, it yields a significant improvement on previous state-of-the-art methods on a variety of challenging benchmarks. Furthermore, it is quite efficient and takes less time to train than previous state-of-the-art methods.

\section*{Acknowledgements}
This work was supported in part by a grant from the Israel Science Foundation (ISF), Phenomics - a grant from the Israel Innovation Authority, and by the Gatsby Charitable Foundations.

\bibliographystyle{plain}
\bibliography{root}

\end{document}